\documentclass[letterpaper, 10 pt, conference]{ieeeconf}  

\IEEEoverridecommandlockouts                              

\usepackage{amsmath}    
\usepackage{mathtools}    
\usepackage{amsfonts}
\usepackage{graphicx}   
\usepackage{subcaption}
\usepackage{epsfig} 
\usepackage{dsfont}
\usepackage{color}
\usepackage[normalem]{ulem}
\usepackage{graphics} 
\usepackage{mathptmx}

\usepackage{array}
\usepackage{cancel}
\usepackage{amssymb}
\usepackage{booktabs}

\usepackage{bbm}
\usepackage{lipsum}
\usepackage[ruled,vlined,noend,titlenotnumbered]{algorithm2e} 
\usepackage[textsize=tiny]{todonotes}

\title{\LARGE \bf
Learning Generalizable Locomotion Skills with\\ Hierarchical Reinforcement Learning 
}

\author{Tianyu Li$^{1}$, Nathan Lambert$^{2}$, Roberto Calandra$^{1}$, Franziska Meier$^{1}$, Akshara Rai$^{1}$
 \thanks{$^{1}$Facebook, Menlo Park, CA, USA, $^2$Work done during an internship at Facebook AI Research, Department of Electrical Engineering and Computer Sciences, University of California, Berkeley, USA
        {\tt\small \{tianyul,rcalandra,fmeier,akshararai\}@fb.com, nol@berkeley.edu}%
}}

\begin{document}

\maketitle
\thispagestyle{empty}
\pagestyle{empty}

\begin{abstract}
Learning to locomote to arbitrary goals on hardware remains a challenging problem for reinforcement learning. 
In this paper, we present a hierarchical learning framework that improves sample-efficiency and generalizability of locomotion skills on real-world robots. Our approach divides the problem of goal-oriented locomotion into two sub-problems: learning diverse primitives skills, and using model-based planning to sequence these skills. We parametrize our primitives as cyclic movements, improving sample-efficiency of learning on a 18 degrees of freedom robot. Then, we learn coarse dynamics models over primitive cycles and use them in a model predictive control framework. This allows us to learn to walk to arbitrary goals up to $12$m away, after about two hours of training from scratch on hardware. Our results on a Daisy hexapod hardware and simulation demonstrate the efficacy of our approach at reaching distant targets, in different environments and with sensory noise.

\end{abstract}

\section{INTRODUCTION}
Reinforcement Learning (RL) can help robots generalize to unseen scenarios, and achieve novel tasks. In locomotion, there has been recent success in using RL to learn to walk in simulation \cite{peng2017deeploco, frans2017meta, chua2018deep}, but examples of RL on locomotion hardware are rare. This is due to multiple reasons, such as sample inefficiency of RL methods, lack of robust locomotion platforms, challenging dynamics, and high-dimensional robots. However, locomotion skills are important for autonomous agents to accomplish tasks outside of their workspace, such as clean a room, pick a far-away object. For navigating uneven terrains, stairs, etc. legged platforms become important. 


In this work, we address two of the main challenges facing learning for legged locomotion research - sample efficiency and generalization. Typical examples of RL on locomotion platforms involve learning conservative policies in simulation and deploying them on hardware \cite{tan2018sim, li2019using, hwangbo2019learning}. 
However, the learned policy might not be efficient on hardware, and might frequently fail \cite{li2019using}. This motivates learning directly on hardware. \cite{haarnoja2018learning} were successful in learning to walk on a Minitaur robot, but training on a higher degree of freedom robot can be very expensive, and most locomotion platforms cannot withstand such extended use. Moreover, \cite{haarnoja2018learning} do not generalize to targets other than walking forward. 


In fact, most works on RL for locomotion try to learn to walk forward, but, any realistic task for locomotion would involve reaching a particular goal in space, in the shortest amount of time or with minimum energy. There is a surprising lack of learning literature that address the problem of reaching arbitrary goals, while there are multiple optimal control papers that address this \cite{feng2016online, mason2018mpc}. This is because generalizing to unseen, arbitrary goals often requires a dynamics model. However, many optimal control algorithms are also sensitive to modeling inaccuracies in dynamics and their performance can suffer with poor models \cite{feng2016online}. 
\begin{figure}[t]
	\centering
    	\includegraphics[width=\linewidth]{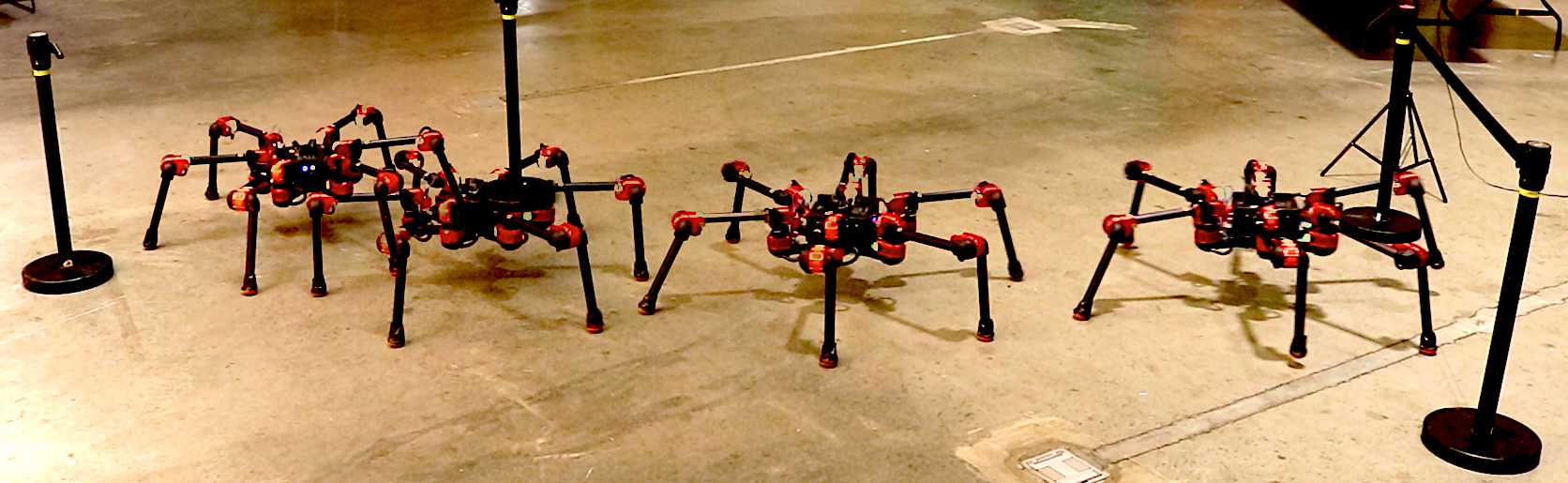}
    \caption{The hexapod \textit{Daisy} used in the experiments. Using our hierarchical control framework, Daisy learned from scratch to reach goals as far as 12 meters in 2 hours of training.}
    \vspace{-0.5cm}
    \label{fig:daisy}
\end{figure}

Model-free learning methods like \cite{andrychowicz2017hindsight} can generalize to goals in the space explored during learning, but do not generalize well to arbitrary goals. Model-based RL holds the promise of generalizing to new goals, but it is largely validated in simulation~\cite{Yang2018Learning}. \cite{bechtle2019curious} point out that learning dynamics models in the model-based RL loop is challenging and might need specialized exploration. As a result, there is little to no evidence of learning to reach arbitrary goals in locomotion literature on hardware.

 In this work, we improve sample-efficiency of RL on locomotion by using a cyclic parametrization of walking policies, similar to \cite{crespi2008controlling, owaki2017quadruped,Yang2018Learning}. We learn the parameters of these policies using a model-free RL algorithm, Soft Actor Critic \cite{haarnoja2018soft}, from scratch on a 18 degree of freedom hexapod robot.
 This cyclic structure is capable of achieving many different locomotion behaviors and gaits without expert intervention, as demonstrated in our experiments. 
 
 Further, we improve generalization to multiple goals by proposing an efficient hierarchical structure. We divide the problem of goal-oriented locomotion into two sub-problems: first we learn temporally extended action primitives that can achieve simple goals such as turning and walking straight, using model-free RL. Next, we build `coarse' dynamics models of these primitives and use them for planning using model predictive control. Coarse dynamics models are fit over transitions over one cycle of primitive actions. This allows us to build dynamics models with very small amount of hardware data, and plan efficiently in primitive space. An overview of our algorithm is shown in Figure~\ref{fig:control_flow_chart}. 


Our main contribution is a hierarchical framework that combines model-free learning with model-based planning to improves generalization of locomotion skills to new goals. Our approach is easy to train, and robust to hardware noise. 
We demonstrate our results on a Daisy hexapod (Figure~\ref{fig:daisy}) over multiple targets up to $12m$ away, starting with training on very short episodes. To the best of our knowledge, this is the first demonstration of such a hybrid model-free learning with model-based planning framework on a locomotion robot hardware. Our results show that such a combination of the two approaches can greatly improve the sample-efficiency and generalization abilities of RL methods for locomotion.

\section{Background and Related Work}

Here, we present a brief overview of model-based and model-free optimization methods from literature and previous works that are closely related to our work.

\subsection{Model-based and model-free optimization}
We consider a Markov Decision Process with actions $a$
, state $s$
and dynamics governed by transition function $f_\theta$.
Starting from an initial state $s_0$, and sampling an action $a_t$ at state $s_t$ according to policy $\pi$, the agent gets a reward $r(s_t,a_t)$  and transition to the next state $s_{t+1} = f_\theta(s_t, a_t)$, generating a trajectory $ \tau = \{s_0, a_0, s_1, a_1, \cdots \}$.

In planning and control, the objective is to maximize the cumulative reward $J_{\pi} = \sum_{t=0}^{T} \mathbb{E}_{\tau_\pi}[r(s_t,a_t)]$, where, $\tau_\pi$ denotes the trajectory distribution generated by policy $\pi$. This can be done in a model-free manner \cite{sutton2018reinforcement} or using information from the dynamics $f_\theta$ in a model-based way.  

\subsubsection{Model-free reinforcement learning}
Model-free RL optimizes a policy $\pi$ by directly maximizing the long-term reward, without reasoning about the dynamics. Model-free methods are typically less sample-efficient that model-based but achieve better asymptotic performance. Our model-free learning algorithm of choice is Soft Actor Critic (SAC) \cite{haarnoja2018soft}, which is a maximum entropy RL algorithm that maximizes both the long-term reward and the entropy of the policy. For a finite horizon MDP, the SAC objective function is:
$J_\pi= \sum_{t=0}^{T} \mathbb{E}_{\tau_\pi}[r(s_t,a_t) - \alpha_tlog\pi(a_t|s_t)]$ where, $\alpha$ is the temperature that trades off between reward and entropy.

SAC is an off-policy algorithm that allows using past data to update current policy, improving sample-efficiency. 
It has been demonstrated to work on locomotion hardware from scratch in \cite{haarnoja2018learning}, and hence we decided to use it. 

\subsubsection{Model Predictive Control (MPC)}

An alternative to model-free RL is to utilize the dynamics (if known) to maximize the long term reward of a trajectory. One such popular approach is MPC, also known as receding horizon control. MPC solves for an action sequence $a_{0:T}$ that maximizes the long-term reward $J = \sum_{t=0}^{T} \mathbb{E}[r(s_t,a_t)]$ subject to the dynamics $s_{t+1} = f_\theta(s_t, a_t)$ at each instant\cite{mayne2000constrained}. The first action $a_0$ is applied on the system, and the process repeats. 

MPC has been widely used for control of dynamical systems, \cite{di2018dynamic, park2015online, herdt2010online}. \cite{mason2018mpc, koenemann2015whole} use MPC for controlling a humanoid robot's center of mass dynamics. However, these works assume a known dynamics model and are sensitive to dynamics modeling errors. As a result, they are hard to generalize to new tasks or robots.

\subsubsection{Hierarchical RL (HRL) with primitives}
Using a hierarchical structure that decomposes complex task control into easier sub-tasks control can speed up learning \cite{sutton1999between}. Previous work studied learning the different levels of the hierarchy together \cite{daniel2016probabilistic, bacon2016optioncritic, stulp2011hierarchical}. An alternative is to divide the task into learning primitives, followed by planning in primitive space, while fine-tuning primitives \cite{stulp2011hierarchical, daniel2013learning, 2017-TOG-deepLoco, frans2017meta}. However, most HRL literature is model-free and hence sample inefficient. For example, \cite{peng2017deeploco} needs over a million samples to learn high-level control. 

We combine model-based planning and model-free learning, by using model-free RL for learning action primitives, and sequencing them using model-based planning. By incorporating dynamics models in HRL, we can improve sample-efficiency as well as generalization.

\begin{figure}[t]
	\centering
	\begin{subfigure}[b]{0.4\textwidth}
    	\includegraphics[width=\textwidth]{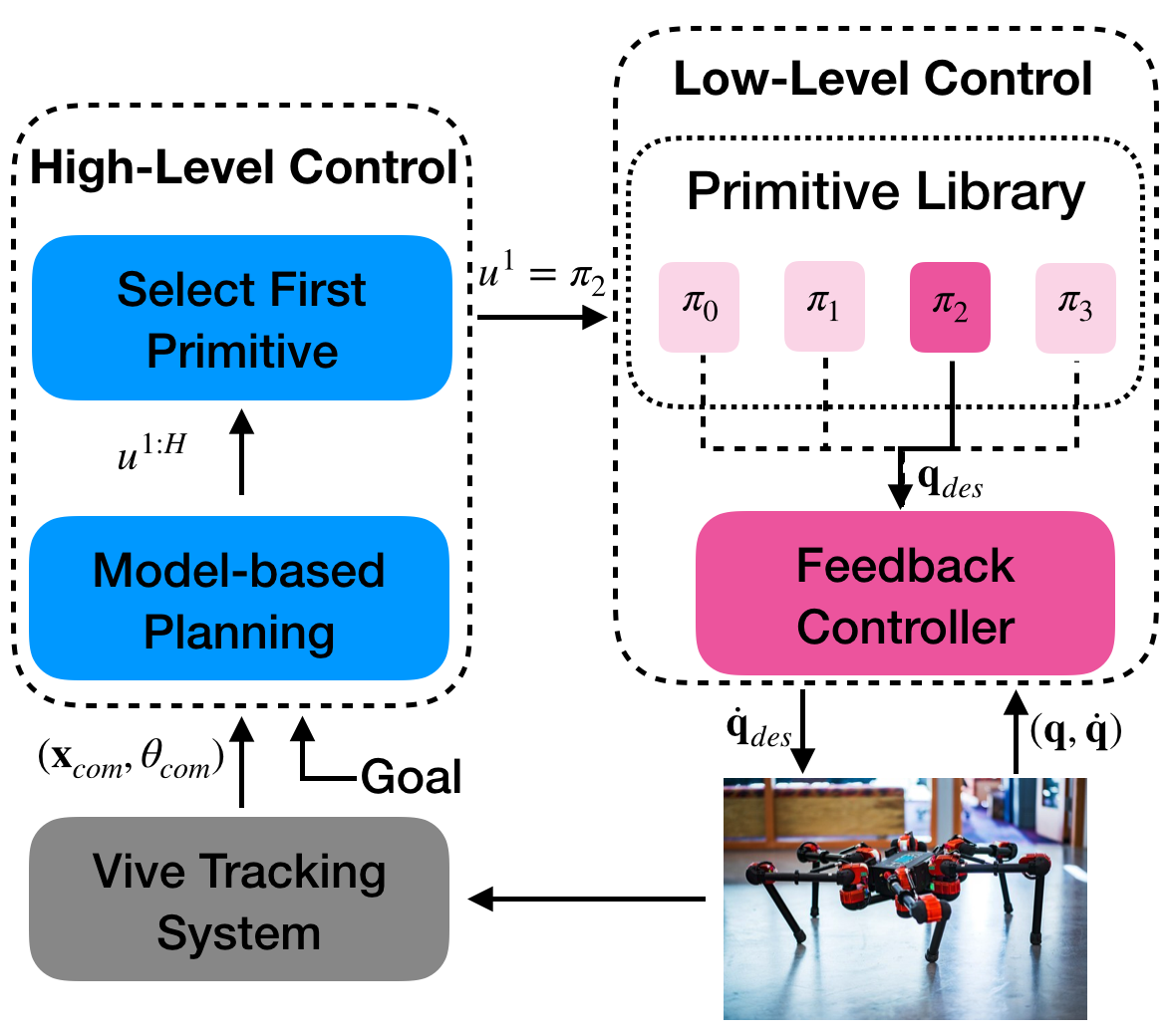}
    \end{subfigure}
    \caption{Hierarchical Control Flow Chart}
    \label{fig:control_flow_chart}
    \vspace{-0.4cm}
\end{figure}

\subsection{Learning for locomotion}

Using Deep RL in locomotion system has been wildly studied in simulation. \cite{peng2017deeploco, MCPPeng19} used hierarchical RL to achieve challenging locomotion tasks in simulation such as moving a soccer ball and carrying an object to a goal. \cite{heess2017emergence} used deep RL to train locomotion systems in different training environments, and found new emergent gaits. \cite{2016-TOG-deepRL} show that robust locomotion can even be achieved from high-dimensional inputs, such as images. However, since these methods take millions of samples, they are not usable on hardware without modifications.

 \cite{Yang2018Learning, antonova2019bayesian, andre2015adapting, cully2015robots} use a cyclic controller structure similar to ours, and use model-free policy search approaches to learn locomotion skills. 
 These methods effectively transfer information between different tasks, or from simulation and hardware. However, they are limited to a relatively low-dimensional parametric controller, and can be hard to generalize to new robots.  On the other hand, \cite{tan2018sim, li2019using} used Deep RL to train unstructured neural network policies in simulation and transfer them to hardware successfully.
However, policies that perform well in simulation do not perform well on hardware due to differences between simulation and the real world, such as contact models. 

Instead of training in simulation and transferring policies to the real-world, \cite{haarnoja2018learning, yang2019data} directly trained policies in the real world on a Minitaur robot. \cite{haarnoja2018learning} used SAC with automatic entropy adjustment to train a Minitaur robot to walk forward with 0.1 million samples. \cite{yang2019data} used model-based RL with trajectory generators to train Minitaur robot to walk in 45,000 samples. 
Minitaur has 8 motors that control its longitudinal motion, and no control for lateral movements. In comparison, our hexapod (Daisy) has omni-directional movements and 18 motors. This makes the problem of controlling Daisy especially challenging, and would require significantly longer training. Moreover, previous work only learns to walk forward, and needs additional training to achieve new goals. Our approach can learn to control Daisy and achieve arbitrary goals, using only 35,000 samples on hardware. Such reduction in hardware samples is important as most locomotion robots get damaged from wear and tear when operated for long. For example, in the course of our experiments, we had to replace two motors.

\section{Learning  Generalizable  Locomotion  Skills}

We now describe our proposed approach in detail. Figure~\ref{fig:control_flow_chart} shows an overview of the hierarchical control structure proposed in this work. In a nutshell, our approach builds a library of primitives $\mathcal{L}= (\pi_0, \pi_1, \pi_2, \pi_3)$ that encode low-level controllers for $4$ micro-actions \emph{turn left}, \emph{turn right}, \emph{move forward}, \emph{stand still}. These primitives are learned via model-free reinforcement learning. On a higher level, our approach depends on a model $f$ that predicts the dynamics of applying a cycle of the primitive. A model-predictive planner utilizes this model to optimize for the next optimal action sequence to achieve a goal. In the following we start by introducing notation and our experimental platform, we then propose two representations for the action primitives and how to learn them and finally describe the high-level planner.


\subsection{Daisy - Hexapod}
Our test platform in this paper is the Daisy Hexapod (Figure \ref{fig:daisy}). Daisy is a six-legged robot with three motors on each leg - base, shoulder, and elbow. 
Practically, the robot is omni-directional, and the center of mass can follow any given trajectory, but the mass of the motors limits the leg velocity. A Vive tracking system is used to measure robot's position in the global frame.

The robot has 18 motors that we control by sending desired motor velocities as actions $a$. The state $s$ used in the high-level planner is the center of mass position and orientation. The low-level policies output 18 desired joint angles $\mathbf{q_{des}}$, which are then converted into desired motor velocities in a feedback loop: $\mathbf{\dot{q}_{des}} = k_p(\mathbf{q_{des}} - \mathbf{q}) - k_d  \mathbf{\dot{q}} + \mathbf{\dot{q}_{ff}}$.
Here $\mathbf{q}$ are the current joint angles; $\mathbf{\dot{q}_{ff}}$ is a feedforward velocity. 
$k_p$ and $k_d$ are hand-tuned feedback gains.


\subsection{Action Primitive Representations}
\label{sec:low-level-control}
We take inspiration from biological gaits in locomotion and use two cyclic parametrizations for our action primitives $\pi_i$. Our primitives take as input a phase variable $t \in (0,1]$ and predicts the next desired joint configuration as an action, $\mathbf{q_\text{des}} = \pi_i(t)$. 
At the beginning of every cycle, the phase variable is initialized to $0$, and then grows linearly to $1$, with the length of the cycle designed by the expert. This also allows us to change the speed of our primitive, for example when training we use a slower primitive for the safety of our robot, but when testing, we increase the frequency for better performance. 

This idea of periodic gaits was also used in \cite{fukuoka2013analysis,kimura2006biologically, owaki2017quadruped}, but these works designed the primitives manually. Instead, here we consider parametric policies, and learn the parameters using a modified SAC, described in Section \ref{sec:SAC}. We consider 2 types of parametrizations for our primitives:
\subsubsection{Neural Network Policy}
An unstructured neural network controller. The input to this network is the 1-dimensional phase $t$ and the output are the 18 desired joint angles. The neural network consists of 2 hidden layers with 64 nodes and a Relu activation function. We also add tanh to the output layer to saturate the outputs.

\subsubsection{Sinusoidal Policy}

A structured parametric controller, which consists of sine waves in each joint: $\mathbf{q^{i}_{des}}(t) =  \mathbf{A_{i}} \sin (2\pi t + \mathbf{B_{i}}) + \mathbf{C_{i}}$. Each motor $j$ has an independent phase $B_{ij}$, offset $C_{ij}$ and amplitude $A_{ij}$ for the $i-th$ primitive, leading to a total 54 dimensional controller $\pi_i$. The parameters of this controller are also learned using modified SAC.

\subsection{Soft Actor-Critic with KL Constraint}
\label{sec:SAC}
While maximum entropy in SAC makes the learning on hardware robust and sample-efficient, sometimes it leads to aggressive policy updates that might harm our robot. Hence, we add an additional practical constraint. We introduce a KL divergence constraint from the previous policy, similar to Trust region policy optimization (TRPO) \cite{schulman2015trust}. Now the objective function for updating policy is expressed as:
$J_\pi = \sum_{t=0}^{T} \mathbb{E}_{\tau_\pi}[r(s_t,a_t) - \alpha_tlog\pi(a_t|s_t)] + \epsilon \ D_{KL}(\pi(\cdot | s)||\pi_{old}(\cdot | s)) $. This cost encourages entropy-based exploration, while keeping the updated policy close to the last policy, leading to more conservative policy updates that still explore.

\subsection{High-Level Control: Model-based Control}\label{sec:high-level-control}
We use a model-based high-level planner that plans the best primitive sequence $u^{1:H}$ for our horizon $H$ using MPC. The dynamics used in this planning are learned over the whole primitive cycle, rather than the instantaneous dynamics, i.e, $s^i_{t+T} = f_\theta(s_t, \pi_i)$ is the next state after executing the primitive $\pi_i$ for $T$ time steps, starting from $s_t$.  This leads to a `coarse' transition model learned over extended action sequences, rather than per time step transitions. Moreover, the planning is in a much reduced space of primitive actions instead of the whole action space.

Starting from the current center of mass position and orientation $(\mathbf{x}_{com}, \mathbf{\theta}_{com})$, our high level planner does an exhaustive search over the possible sequences of actions to find the globally optimal sequence for our horizon $H=3$. Moreover, to further simplify the dynamics, we learn a delta dynamics model $\delta s = f_\theta(\pi_i)$, which reasons about the change in the state after the execution of the primitive. This makes the dynamics learning much more efficient, and generalize to unseen states.


\begin{algorithm}[t]
\SetAlgoLined
 Define primitives $\pi_{1:K}$ and rewards $r_{1:K}$ \\
 \For{each primitive}{

  \For{each environment step}{
       $a_t \thicksim \pi_i(a_t|s_t)$ \\
       Apply action $a_t$, measure $s_{t+1}$\\
       $D \leftarrow D \cup \{(s_t, a_t, s_{t+1},r^{1:K}_t)\}$
   }
  \For{each gradient step}{
    Update $Q_i$, $\pi_i$\\
  }
}
Given primitive library $\mathcal{L} = \{\pi_1, \cdots \pi_{K}\}$, cycle time $T$\\
\For{each dynamics learning step}{
\For{each primitive $\pi_i$}{
       $a_t \thicksim \pi_i(a_t|s_t)$ \\
       Apply action $a_t$, measure $s_{t+1}$\\
   }
   $D' \leftarrow D' \cup \{(s_0, \pi_i, s_{T})\}$\\
   Learn dynamics model  $s_{T} = f_\theta(s_0,\pi_i)$ \\
   $s_0 \leftarrow s_{T}$\\
}
Given reward $r_{hl}$, primitive library $\mathcal{L}$, horizon $H$, $s_0$\\
\For{each planning step}{
    $u_{1:H} = \arg \min r_{hl}(s_0, u_{1:H})$ \\
    Apply primitive $u_1$, measure $s_T$ \\
    $s_0 \leftarrow s_T$
}

 \caption{Hierarchical Reinforcement Learning }
 \label{algo:HRL algo}
\end{algorithm}

\section{Experimental Results}
In the following we present evaluations on the Daisy hexapod both in simulation and hardware. Our experiments are structured as follows:
\begin{itemize}
    \item Learning of primitives: We train two primitive actions on Daisy : walk forward, and turn. During training, the total steps in a cycle is 100, and we sample 10 cycles for each iteration, hence 1000 samples per iteration.
    \item High-level control: For experiments with the high-level control, we use MPC for planning in the space of trained primitives. We set targets far away from the training region of the primitives, and reach them using the hierarchy.
\end{itemize}

\subsection{Simulation Experiments}
We simulate the Daisy robot in PyBullet~\cite{pybullet}. We start by describing our experimental setup for learning the low-level primitives in simulation.
\subsubsection{Learning Primitives}
We decompose locomotion behaviors into four elementary motions: \emph{move forward}, \emph{turn left}, \emph{turn right}, and \emph{stand-still}. Since turning right can be achieved by mirroring the control of turning left, we do not need to train a new policy; for standing still the desired joint state is the current joint state.

We train the \emph{move forward} and \emph{turn right} primitives in sequence, starting with \emph{move forward}. The parameters of the \emph{move forward} policy are initialized randomly, and the training data is used to initialize training of \emph{turn right} policy. In simulation both primitives are trained for 50 iterations using the algorithm described in Section \ref{sec:SAC}. 

For training the \emph{move forward} policy, we used the reward function
\vspace{-0.3cm}
\begin{equation}
    r_t = \pmb{w}_1\delta \pmb{x}_{com,t} -\pmb{w}_2|\pmb{\theta}_{com,t} | - \pmb{w}_3|\Dot{\pmb{q}}_{joint,t}|\,,
\end{equation}
where the first term 
gives reward for moving forward and penalty for lateral and backward movements, the second term tries to minimize deviation in orientation, and the third penalizes for high joint velocities. 

After training the \emph{move forward} policy, we switch to training the \emph{turn right} policy. We reuse the data collected in this the phase to initialize the parameters of the \emph{turn right} policy. Since SAC is an off-policy method, we can just re-evaluate the reward of each transition on the \emph{turn right} reward function and restart training. The reward function to train the \emph{turn} policy was
\vspace{-0.2cm}
\begin{equation}
    r_t = -\pmb{w}_1|\delta \pmb{x}_{com,t}| -\pmb{w}_2|\pmb{\theta}_{com,t}  -  \pmb{\theta}_{des,t} |  -  \pmb{w}_3|\Dot{\pmb{q}}_{joint,t}|\,.
\end{equation}
This reward function penalizes the movement of the center of mass in any direction. For each primitive cycle, we assign a desired orientation for the robot. Lastly, we penalize high joint velocity for the safety of our robots. Intuitively, this reward functions encodes that the optimal turning behavior is to turn on the spot at a constant speed. The parameters for reward functions for training are shown in Table~\ref{tab:reward_fun}.
\begin{table} [t]
\begin{center}
\begin{tabular}{ l c c c  } 
 \hline
 $\ $ & $\mathbf{w}_1$ & $\mathbf{w}_2$ & $\mathbf{w}_3$\\ 
  \hline
 sim-forward & [-1,5,-0.1] & [0.1,0.1,0.1] & [0.01]\\ 
 sim-turn & [0.1,0.1,0.1] & [0.1,0.1,20] &  [0.002]\\
 hw-forward & [-50,300,-10] & [1,1,1] & [0.01]\\
 hw-turn & [1,1,1] & [0.1,0.1,40]  & [0.002]\\
 \bottomrule
\end{tabular}
\caption{Low-level reward function weights}\label{tab:reward_fun}
\vspace{-0.5cm}
\end{center}
\end{table}

\begin{figure*}[t]
    \begin{subfigure}[b]{0.245\linewidth}
    	\includegraphics[width=\textwidth]{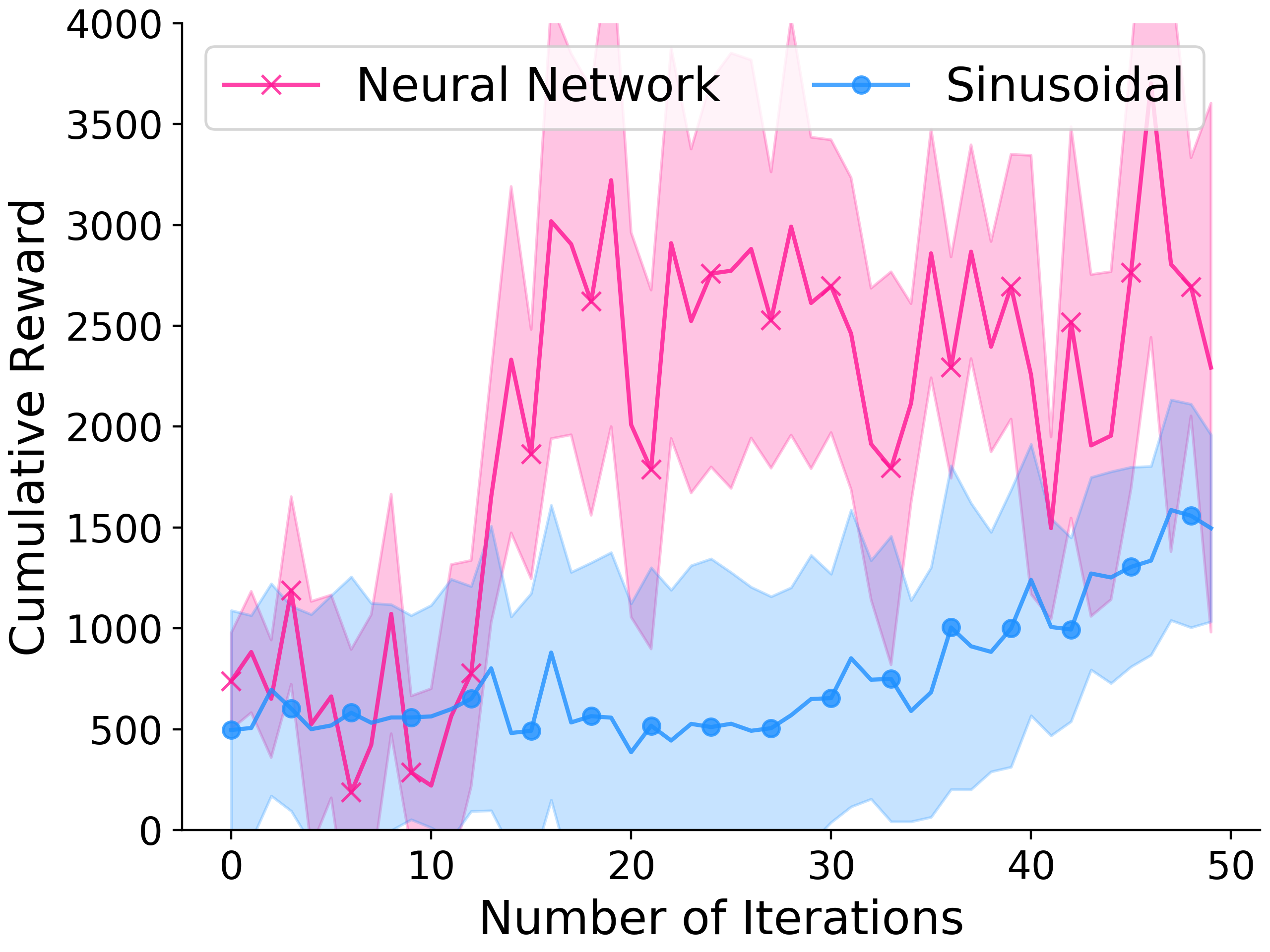}
        \caption{Simulation: Forward}
        \label{fig:sim_low_train1}
    \end{subfigure} %
    \hfill
    \begin{subfigure}[b]{0.245\linewidth}
    	\includegraphics[width=\textwidth]{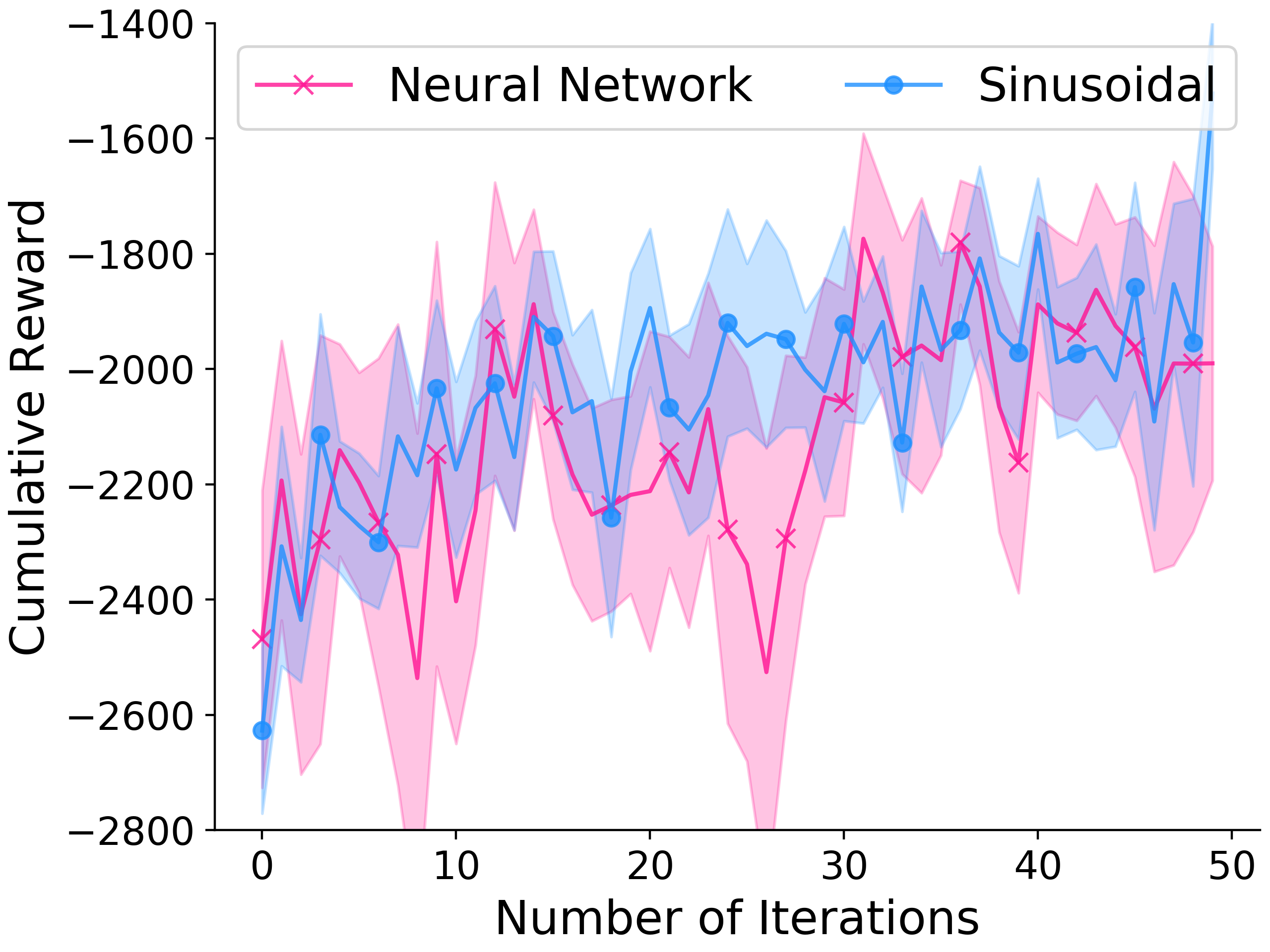}
        \caption{Simulation: Turning}
        \label{fig:sim_low_train2}
    \end{subfigure}%
    \hfill
	\begin{subfigure}[b]{0.245\linewidth}
    	\includegraphics[width=\textwidth]{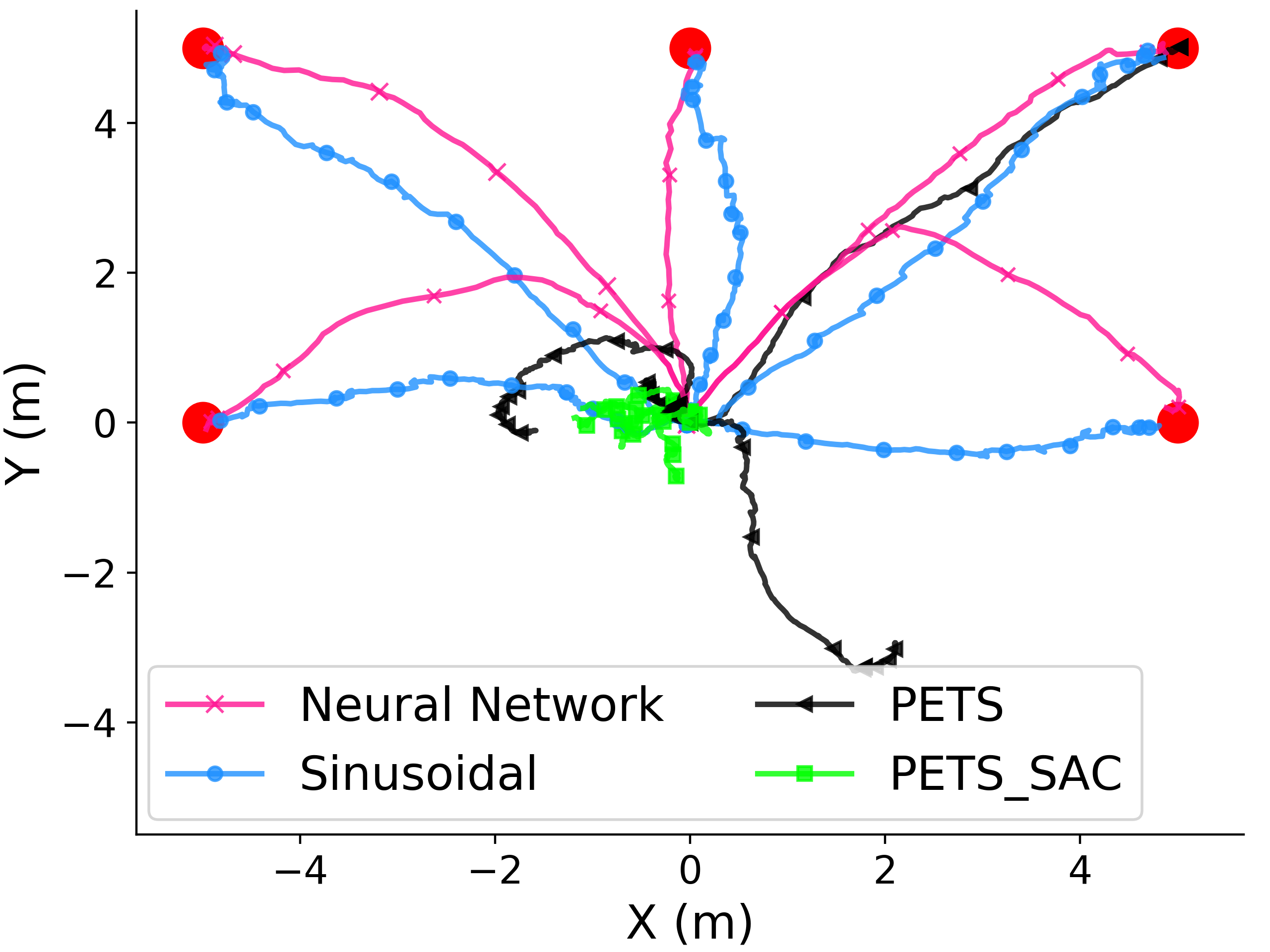}
    	\caption{Different Goals}
    	\label{fig:sim_many_goals}
    \end{subfigure} %
    \hfill
    \begin{subfigure}[b]{0.245\linewidth}
    	\includegraphics[width=\textwidth]{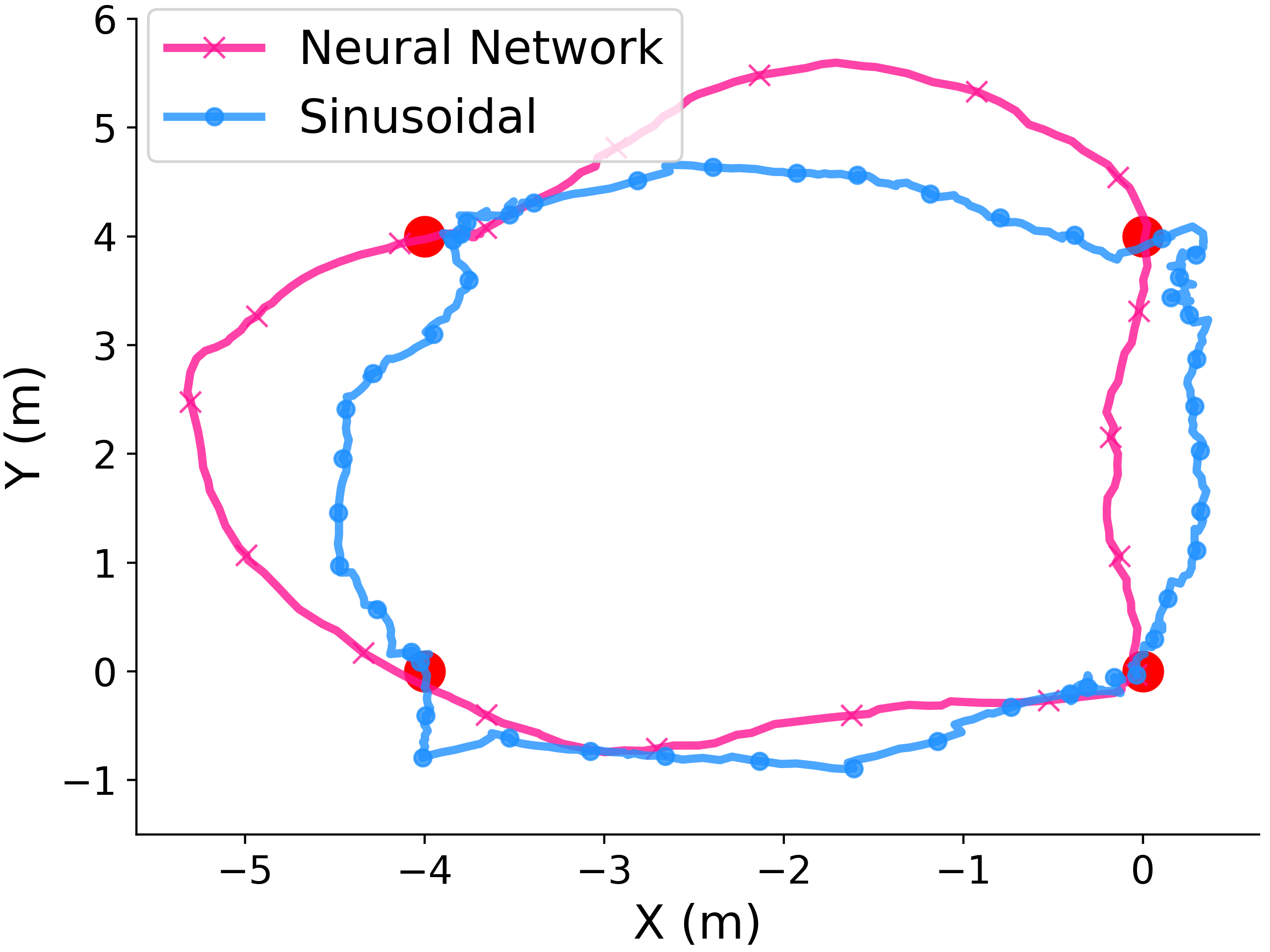}
        \caption{Waypoint Goals}
        \label{fig:sim_square}
    \end{subfigure}
    \caption{(a,b) Simulation training plot for forward turning controllers. We collect 1000 samples per iteration.  (c,d) Simulation experimental results of Daisy reaching different goals. (c) Comparison of our approach vs. PETS for achieving different goals starting from $(0,0)$. (d) Daisy moving to the corners of a square starting from $(0,0)$.}
    \vspace{-0.6cm}
\end{figure*}

%
%
Our simulation training results for the neural network and sinusoidal controller are shown in Figure \ref{fig:sim_low_train1}, \ref{fig:sim_low_train2}. In simulation, for the forward task, the neural network  learns faster than the sinusoidal controller, and the reward is also higher than the sinusoidal controller. This is because the ground contact models in the simulation are very inaccurate, and with the neural network controller, the optimization quickly learns to exploit it by sliding. Since turning is a more controlled task with a target orientation, it is harder to exploit the simulation and both controllers learn at a comparable rate, with the sinusoidal controller having a more stable learning curve.


\subsubsection{High-level control}

Once the primitive actions are trained, we can move to the high-level control. We start by training a dynamics model for each primitive by simply building a look-up table for $\delta \mathbf{s} = f(\pi_i)$. The look-up table is trained by sampling 50 cycles of randomly selected primitives and averaging the resultant displacement, as described in Algorithm \ref{algo:HRL algo}.

Once the look up table has been created, we utilize the model within MPC to optimize the sequence of actions that minimizes the cost over a horizon of $H=3$. We apply the first action from this optimized sequence on the robot, and replan. The reward for the high-level control $r_{hlc, sim}$ is episodic, the final distance between the robot and goal at the end of the horizon $r_{hlc, sim}=-|\mathbf{x}_{goal}-\mathbf{x}_{com}|$.

In simulation, we compare the high-level control against PETS \cite{chua2018deep}, a state-of-the-art model-based RL baseline. We compare against two versions of PETS:
\begin{itemize}
    \item PETS : We train the full dynamics model of the robot while trying to achieve a goal, in the standard PETS loop. Then we do MPC with cross-entropy method (CEM) using the trained dynamics to achieve other goals, far away from the goal for which the dynamics was trained.
    \item PETS with SAC data : We train the full dynamics model on data that was used for training the forward and turning controllers. This dynamics includes turning and walking data, but for a very small part of the robot's space. The goals are set quite far away from the training set, and MPC+CEM is used to optimize the action.
\end{itemize}
We note that the dynamics trained for PETS comparison are on the full state of the robot (18 joint angles), and the action is an optimized sequence of 18 desired joint velocities. As compared to our hierarchical framework, this is a much higher dimensional optimization problem.

In simulation, we test two experimental settings: 

\begin{enumerate}
    \item \textbf{Different goals:} The goals are at $(5,0)$, $(5,5)$, $(0,5)$, $(-5,5)$, $(-5,0)$ starting from $(0,0)$ (Figure~\ref{fig:sim_many_goals}). Both the neural network and sinusoidal controllers can achieve all targets using our approach. Baselines PETS and PETS trained on SAC data fail to achieve goals other than the one they were trained on.
    \item \textbf{Waypoint goals:} The robot has to achieve targets in a square at $(0,4), (-4,4), (-4,0)$ sequentially, starting from $(0,0)$ (Figure \ref{fig:sim_square}). Both the neural network and sinusoidal controllers can achieve all targets using our hierarchical control structure. This setting is similar to waypoint goals, where the robot sequentially moves between targets. 
\end{enumerate}

In both these experiments (Figure~\ref{fig:sim_many_goals}, \ref{fig:sim_square}), the hierarchical control performs well, and the robot is able to reach the targets, despite slipping in simulation. In comparison, while PETS is able to reach the goal that the dynamics were learned on efficiently, it does not generalize to other goals in the environment. Since PETS with SAC data is only trained on very short episodes, it is also unable to achieve far away goals. Hence, the hierarchy helps improve generalization to new goals, when trained with the same amount of data as PETS, a model-based RL approach.


\begin{figure*}[t]
	\centering
	\begin{subfigure}[b]{0.245\linewidth}
    	\includegraphics[width=\textwidth]{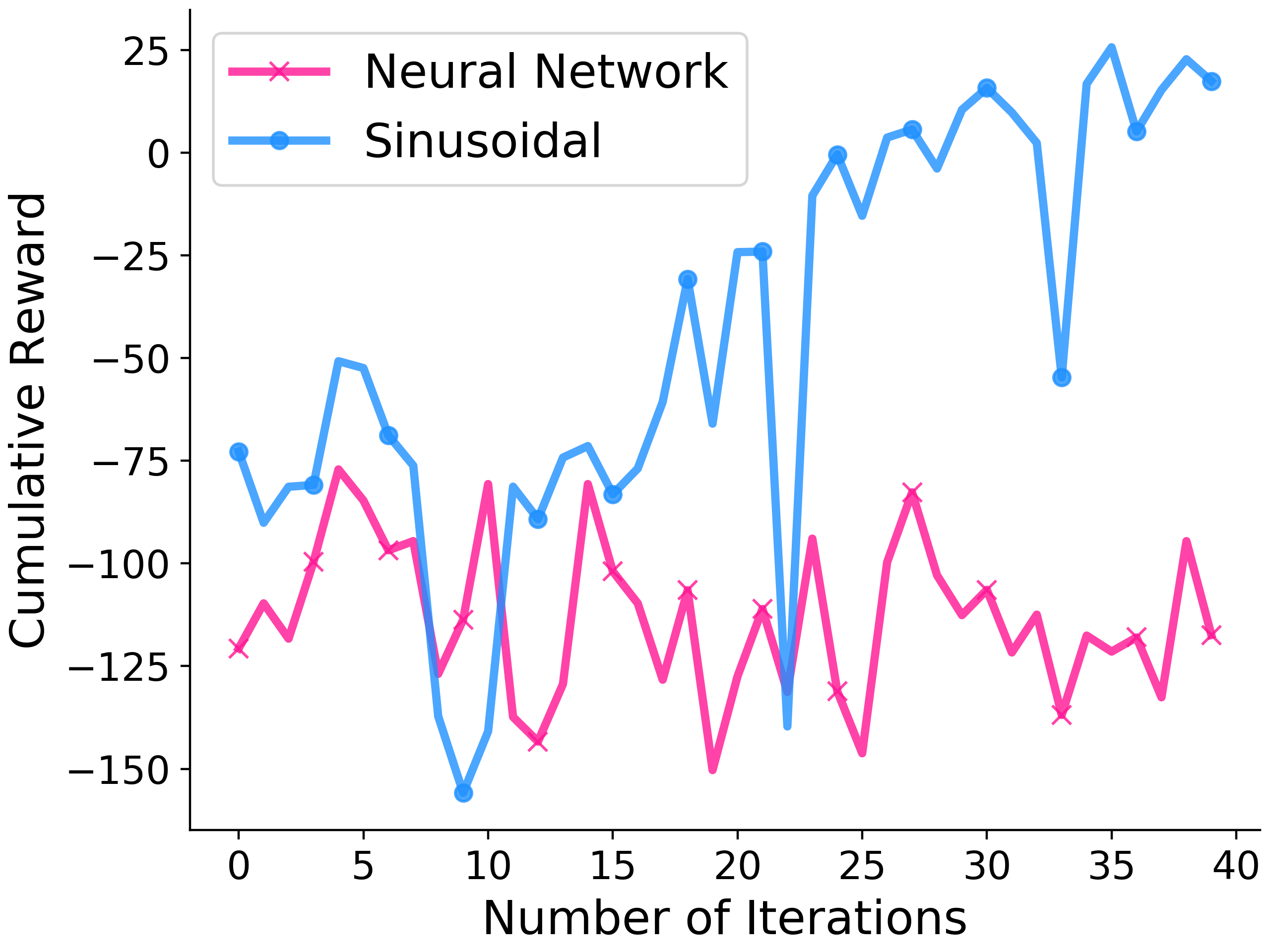}
        \caption{Training: Forward}
    \end{subfigure}%
    \hfill
    \begin{subfigure}[b]{0.245\linewidth}
    	\includegraphics[width=\textwidth]{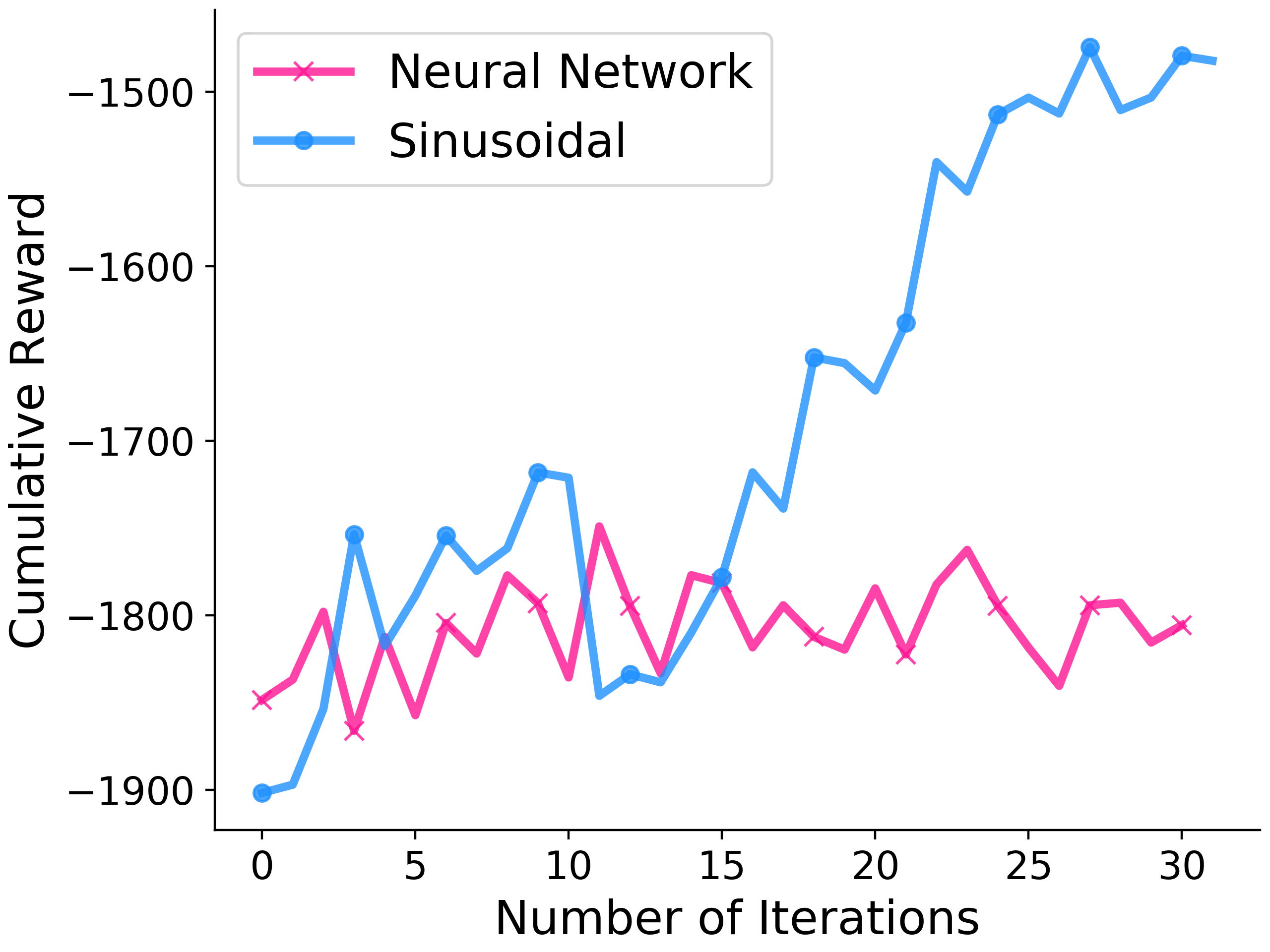}
        \caption{Training: Turning}
    \end{subfigure}%
    \hfill
	\begin{subfigure}[b]{0.245\linewidth}
    	\includegraphics[width=\textwidth]{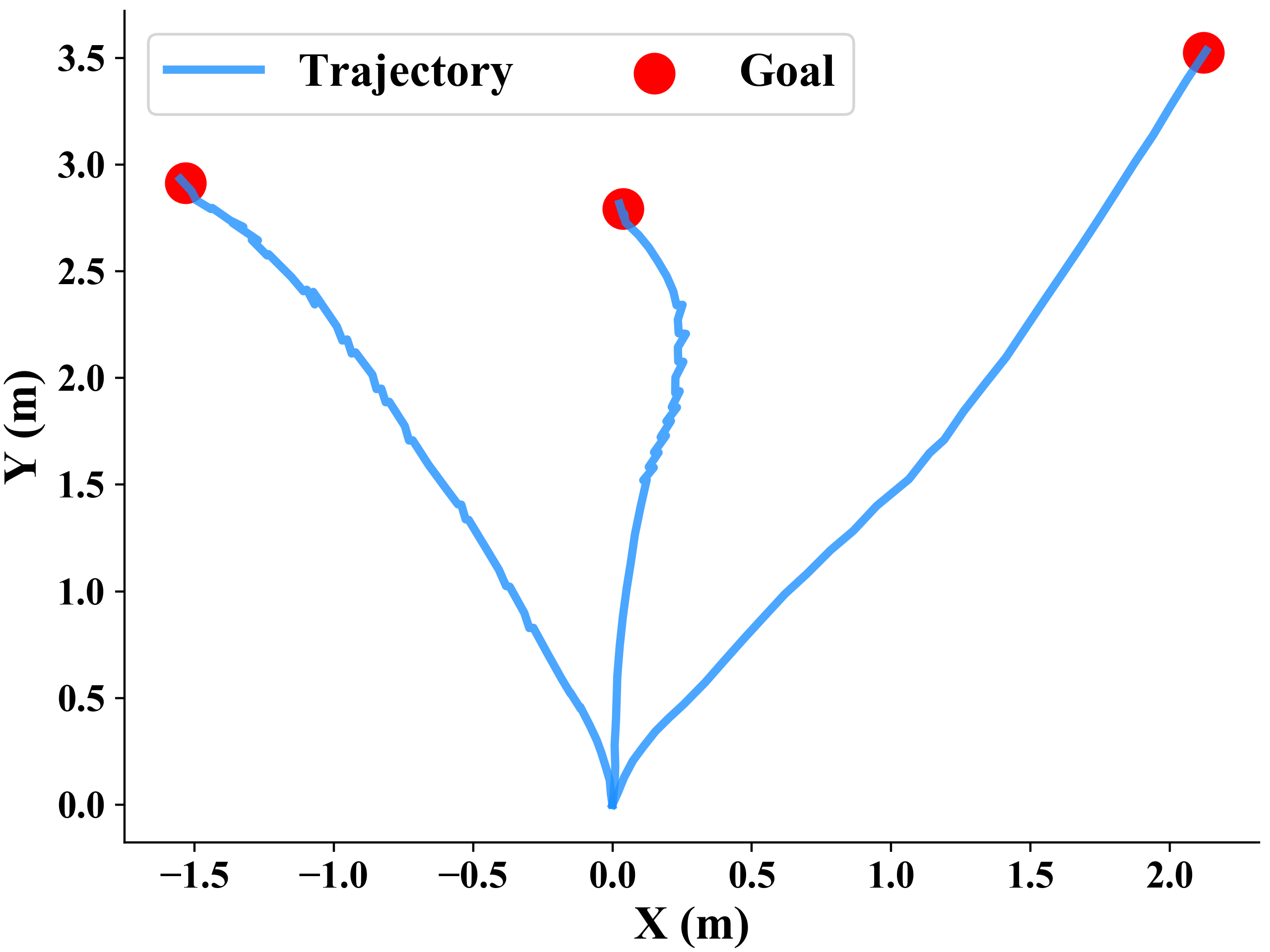}    
    	\caption{Different Goals}
    	\label{fig:hard_many_goals}
    \end{subfigure}%
    \hfill
    \begin{subfigure}[b]{0.245\linewidth}
    	\includegraphics[width=\textwidth]{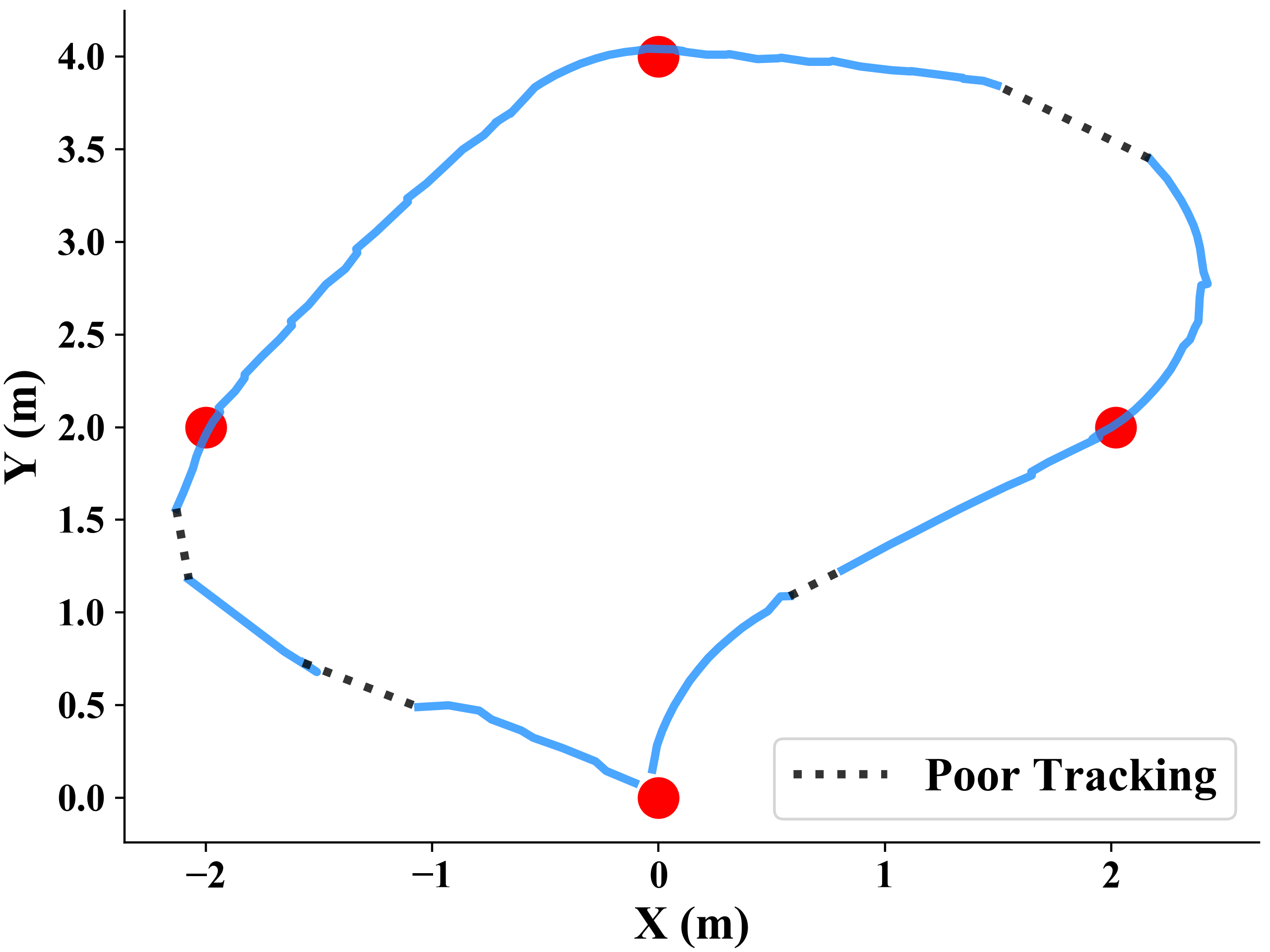}    
        \caption{Waypoint Goals}
        \label{fig:hard_square}
    \end{subfigure}
    \caption{(a,b) Hardware learning curves for forward and turning controller (collecting 500 samples per iteration). (c,d) Hardware experimental results of Daisy reaching different goals. (c) Daisy achieving different goals starting from $(0,0)$. (d) Daisy moving to the corners of a square starting from $(0,0)$. Dotted lines indicate poor tracking.  }
    \label{fig:real_low}
    \vspace{-0.3cm}
\end{figure*}

\begin{figure*}[t]
	\centering
	\includegraphics[width=.94\textwidth]{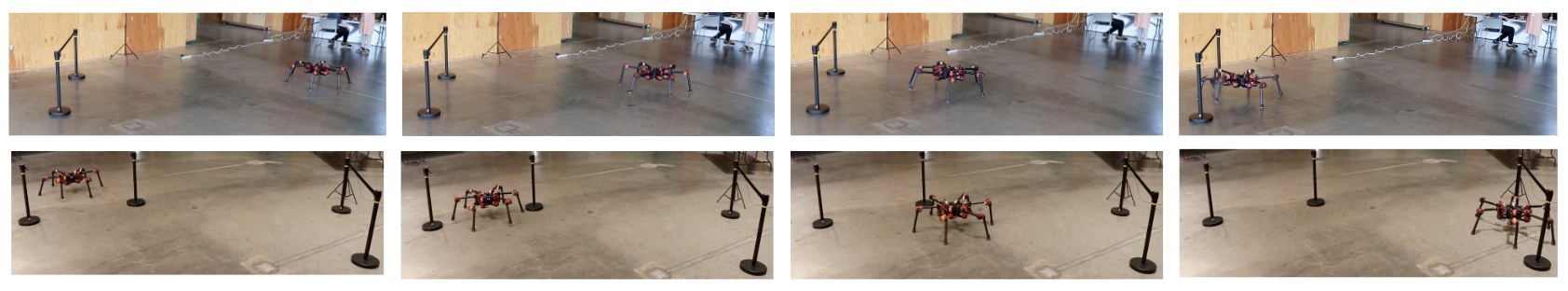}
    \caption{A time lapse of Daisy walking towards different goals.}
    \label{fig:Hardware_walking}
    \vspace{-0.5cm}
\end{figure*}

\subsection{Hardware Experiments}
Simulation experiments allowed us to test the validity of our approach, but did not have an accurate contact model with the ground. The neural network controllers trained in simulation performed very poorly on hardware because of this mismatch, making it essential to train directly on hardware.

\subsubsection{Learning Primitives}
For hardware experiments, we used the same formulation of reward as in simulation but with slightly different weights in rewards, as summarized in Table~\ref{tab:reward_fun}. The parameters of the \emph{move forward} policy are initialized randomly, and the training data is used to initialize training of \emph{turn right} policy. 
%
We trained forward and turning policies on hardware, and their learning curves are shown in Figure \ref{fig:real_low}. We used 20000 samples to train the forward controller which took approximately an hour and 15000 samples to train the turning controller which took about 45 minutes. 
 Although in simulation the neural network trains faster than the sinusoidal controller, we were not successful in training a neural network policy from scratch on hardware, possibly due to noise in reward generation. Since the sinusoidal controller is restricted in space of controllers, in our experience, it was more robust to noise in reward signals, as compared to the neural network controller. The trained sinusoidal forward controller can walk straight and the turning controller can turn left with a small turning radius.

\subsubsection{High-Level Controller}
On hardware, we add an orientation term to the high level reward, because the position sensing tends to drift over time, and the robot fails to reach the global goal without orientation guidance.
\vspace{-0.2cm}
\begin{equation}
r_{hlc, hw}=-|\mathbf{x}_{goal}-\mathbf{x}_{com}| - |<\mathbf{\theta}_{goal},\mathbf{\theta}_{com}>|    
\end{equation}
Here, the first term is the distance term, same as in simulation, and the second term measures the deviation of the center of mass orientation from the goal angle.

For our hardware experiments with high-level control, we start by building the dynamics models of the each primitive in the primitive library $\mathcal{L}$. Each dynamics is trained for 50 samples on hardware, leading to a total of 200 hardware samples. 

For testing our algorithm on the Daisy robot, we designed a similar experimental setup as simulation, where Daisy was commanded to go to goals up to 12m away from its start point. While our method can generalize to arbitrarily far away locations, currently our hardware setup is limited by the sensing of Vive tracking system for global position of Daisy; our goals are limited to be in the region covered by the base stations. Despite this, sometimes the robot loses tracking during the experiments, and the high-level action is hard-coded to stand still until the tracking is recovered. We test two experimental settings on hardware: 

\begin{enumerate}
    \item \textbf{Different goals:} The robot has to move to goals $(-1.5, 3), (0, 3), (2, 3.5)$ starting from $(0,0)$ (Figure \ref{fig:hard_many_goals}). The sinusoidal controller can reliably achieve all targets despite slipping on the ground and noise.
    
    \item \textbf{Waypoint goals:}  The robot is sequentially asked to move to a series of goals. Similar to the simulation setup, the robot has to reach corners in a square sequentially, starting from $(0,0)$ (Figure \ref{fig:hard_square}). Our approach easily generalizes to this setting. In the future, these hand-designed goals can be replaced by a separate controller that specifies waypoints for the robot to move to.
    
\end{enumerate}


Our hardware experiments show that our proposed hierarchical controller can achieve far away goals using very small amount of training data on hardware. It generalizes to different scenarios, as well as different experimental settings like different flooring, sensing noise, etc. Though we could not train the neural network policy successfully on hardware, we achieved reliable success with the sinusoidal policy at reaching far away goals.
However, the performance of the action primitives can be improved on hardware, for example the forward primitive moves at about 0.15m/s. An online updating scheme that fine-tunes the primitives and their dynamics models for a new setting can improve performance on new floors. Moreover, discovering new primitives, such as to go upstairs is a challenging problem. We assume that the size of the primitive library is predefined, but in the future, we would like to explore methods similar to \cite{frans2017meta} to discover new primitives online, while maintaining the generalizability and sample-efficiency of our current approach.

\section{Conclusion}

In this work, we proposed a hierarchical control structure for controlling locomotion robots. We decomposed the problem of learning locomotion skills into two sub-problems -- learning low-level locomotion skills, followed by sequencing them in a model-based way. Our experiments on the Daisy robot show that such a decomposition can lead to very sample-efficient learning of generalizable locomotion skills. Using our approach, Daisy can reach goals up to 12m away from its start location, and follow waypoints defined by a user. 
In the future, these waypoints can be generated by a separate controller that takes the environment state as input, for example with an image.

Our work is a step towards building generalizable locomotion skills that can reach arbitrary goals in unknown environments. However, there are many avenues for improvement over our current performance. The low-level primitives, when trained and tested on different environments can have very different performance. For example, they might slip on a slippery floor, or walk too conservatively. While the high-level control helps achieve targets despite these disturbances, performance can improved by updating the primitives locally for different environments. Additionally, there might be a need to discover new primitives for new settings. For example, if a leg breaks, or in the presence of stairs, the current library of primitives might not be enough to achieve a goal. In such cases, one could try to incrementally learn new primitives for achieving new targets, and store them in the library for future reusability. We leave this to future work.


 \bibliographystyle{IEEEtran}
 \bibliography{bibliography}

\end{document}